\documentclass[11pt]{article}

\usepackage[margin=1in]{geometry}

\usepackage[utf8]{inputenc}
\usepackage[T1]{fontenc}
\usepackage{lmodern}
\usepackage{hyperref}
\usepackage{url}
\usepackage{amsfonts}
\usepackage{amsmath}
\usepackage{amssymb}
\usepackage{mathtools}
\usepackage{microtype}
\usepackage[capitalize,noabbrev]{cleveref}
\usepackage{enumitem}
\usepackage{authblk}
\usepackage{natbib}

\newcommand{\R}{\mathbb{R}}

\newcommand{\In}{\operatorname{In}}

\title{Variational Structure at the Edge of Stability}

\author[1]{Eric Regis}
\affil[1]{Yale University}
\affil[ ]{\texttt{eric.regis@yale.edu}}

\date{}

\begin{document}

\maketitle

\begin{abstract}%
When discrete-time optimizers operate at the edge of stability, they exhibit near-two-periodic behavior. These oscillatory dynamics are reminiscent of conservative systems, such as the dynamics generated by symplectic integrators. However, a precise formulation of the connection between discrete-time optimizers at the edge of stability and discrete mechanics remains underexplored. Recently, Litman introduced the \textit{edge coupling}: a functional on consecutive gradient descent iterates whose critical points encode the fixed points and two-point orbits of the gradient descent dynamics. Here we extend the edge coupling to heavy-ball and Nesterov momentum. We show that its critical points characterize the fixed points and two-point orbits---with its Hessian characterizing their stability. We also show that the edge coupling can be identified with the symmetric Verlet action, formalizing the connection between the edge of stability and discrete mechanics.
\end{abstract}

\section{Introduction}
\label{sec:intro}

Gradient-based training of neural networks typically operates at the edge of stability~\citep{cohen2021gradient}. There, the iterates oscillate along the sharp directions of the Hessian---with the sharpness hovering at a value that depends on the optimizer~\citep{cohen2022adaptive}. The dynamics of EoS are near-two-periodic: after two consecutive updates, the iterates \textit{almost} return to where they started. This is why the loss decreases slowly despite the large steps taken at each update: the iterates bounce across the valley rather than rolling smoothly downhill.

Oscillatory dynamics are more commonly found in conservative systems: a system where energy is conserved instead of dissipated. One could then characterize the dynamics of optimizers like gradient descent and heavy ball momentum at the edge of stability as \textit{near} conservatory. While the parallel is intuitive, formalizing the connection between the dynamics at the edge of stability and those of conservative systems remains underdeveloped.

Recently, \citet{litman2026origin} introduced the \emph{edge coupling}: a functional on consecutive GD iterates whose critical points encode the fixed points and two-period orbits of gradient descent. The edge coupling is reminiscent of objects commonly found in variational mechanics~\citep{marsden2001discrete}. With it, Litman was able to shed light on why the dynamics of gradient descent drive the sharpness to hover at $2/\eta$.

Here we introduce the \textit{phase-space edge coupling}, extending Litman's edge coupling to include heavy-ball momentum. As was the case with gradient descent, its critical points encode the fixed points and two-period orbits---and its Hessian encodes their stability. Furthermore, we show that for all values of the momentum coefficient $\beta$, the phase-space edge couplings are discrete \emph{actions}---specifically, the symmetric Verlet action~\citep{verlet1967computer, hairer2003acta}. In doing so, we have worked to clarify the connection between edge of stability and discrete mechanics.

\subsection{Contributions}
\label{sec:contributions}

Our contributions are:
\begin{itemize}[leftmargin=*,itemsep=0pt,topsep=2pt]
	\item An introduction of the phase-space edge coupling~(\cref{sec:coupling}).
	\item A family of coordinate transformations under which these properties are preserved~(\cref{sec:coordinates}).
	\item A demonstration that the functional's Hessian encodes their linear stability~(\cref{sec:hessian}).
	\item A connection to the symmetric Verlet action and discrete mechanics~(\cref{sec:mechanics}).
\end{itemize}

\section{Related Work}
\label{sec:related}

\textbf{Edge of stability.} \citet{jastrzebski2020breakeven} documented a break-even point on the optimization trajectories of SGD at which the largest eigenvalue of the Hessian stabilizes. \citet{cohen2021gradient} observed that full-batch gradient descent operates at the edge of stability: the sharpness rises until it reaches the stability threshold $2/\eta$ and then hovers there. \citet{cohen2022adaptive} extended the edge of stability to momentum and adaptive methods, whose stability thresholds are modified by the momentum and the preconditioner. 

A body of theoretical work explains the mechanism: \citet{ahn2022understanding} characterize the unstable convergence of gradient descent beyond the classical step-size threshold, \citet{arora2022understanding} prove that in the edge-of-stability regime gradient descent implicitly tracks a sharpness-reduction flow, and \citet{damian2023selfstabilization} identify a self-stabilization mechanism by which growing oscillations feed back to reduce the sharpness. \citet{cohen2025central} derive central flows that capture the time-averaged trajectory of oscillatory optimizers.  \citet{andreyev2026momentum} analyzes heavy-ball momentum at the edge of stochastic stability, and \citet{phunyaphibarn2024gradient} show that Polyak's momentum enlarges the catapults of the unstable phase, biasing the iterates toward flatter minima.

Several works analyze the edge of stability through consecutive iterates. \citet{chen2023beyond} establish convergence of gradient descent beyond the classical stability threshold by analyzing the two-step update map and its period-2 orbits. Rod flow~\citep{regis2026rod} is a continuous-time model in which consecutive iterates at the edge of stability are modeled as an extended object. Its extension to momentum and adaptive methods features a phase-space rod~\citep{regis2026adamrod}. Edge flow~\citep{marion2026edge} is a tractable continuous-time model that predicts the trajectory of gradient descent at the edge of stability. \citet{litman2026origin} introduced the edge coupling on consecutive GD iterates, whose critical points are the fixed points and 2-cycles of gradient descent. This functional is the object the present paper generalizes.

\textbf{Discrete mechanics.} Discrete mechanics derives integrators by discretizing the action rather than the equations of motion. \citet{marsden2001discrete} give a comprehensive treatment of variational integrators and discrete Legendre transforms. The St\"ormer--Verlet (leapfrog) scheme is the canonical example~\citep{verlet1967computer, hairer2003acta}. Symplectic integrators conserve not the true energy but a nearby \emph{shadow} Hamiltonian, a statement made precise by backward error analysis~\citep{hairer2006geometric, benettin1994shadow}. \citet{mackay1983linear} characterize the linear stability of periodic orbits in Lagrangian systems through the Hessian of the discrete action. Heavy-ball momentum itself originates with \citet{polyak1964some}. 

 A separate line of work connects momentum methods to continuous-time variational principles. \citet{wibisono2016variational} derive accelerated methods as discretizations of a Bregman Lagrangian, \citet{franca2020conformal} interpret heavy ball and Nesterov's method as conformal symplectic integrators of dissipative Hamiltonian systems, and \citet{muehlebach2021optimization} argue that momentum methods are best understood through symplectic discretizations of dissipative dynamics rather than through vanishing-step-size limits.

\section{The Phase-Space Edge Coupling}
\label{sec:coupling}

We briefly recap the edge coupling for gradient descent \citep{litman2026origin}. We then introduce its generalization, the phase-space edge coupling.

\subsection{The Edge Coupling for Gradient Descent}
\label{sec:litman}

Recall the update rule for gradient descent:
\begin{equation}
w_{t+1} = w_t - \eta \nabla L(w_t),
\label{eq:gd-update}
\end{equation}
where $w_t \in \R^d$ is the location of the iterates in weight space, $L : \R^d \to \R$ is the loss function, and $\eta>0$ is the step size. 

A property of the loss function is that its critical points are precisely the fixed points of the dynamics of gradient descent.
\begin{equation}
	\nabla L(w_t) = 0 \iff w_{t+1} = w_t.
\end{equation}
The edge coupling is given as:
\begin{equation}
\mathcal{A}_\eta(w, w') = L(w) + L(w') - \frac{1}{2\eta} \|w - w'\|^2.
\label{eq:A-def}
\end{equation}
Let $\nabla \mathcal{A}_\eta$ denote the gradient of the edge coupling with respect to both of its arguments:
\begin{equation}
	\nabla \mathcal{A}_\eta = (\nabla_w, \nabla_{w'}) \mathcal{A}_\eta .
\end{equation}
One interpretation of the edge coupling is that it generalizes the loss function: it characterizes not only the fixed points of gradient descent, but \textit{also} its two-period orbits. Specifically, a pair of consecutive iterates $(w_t, w_{t+1})$ lies on a two-period orbit of GD if and only if $\nabla \mathcal{A}_\eta$ vanishes there:
\begin{equation}
	\nabla \mathcal{A}_\eta(w_t, w_{t+1}) = 0 \iff w_{t+2} = w_t .
\end{equation} 
This can easily be shown. If you take the partial derivative of $\mathcal{A}_\eta$ with respect to $w$ and set it equal to zero, you recover the GD update equation from $w$ to $w'$:
\begin{equation*}
w' = w - \eta \nabla L(w).
\end{equation*}
Similarly, taking the partial derivative of $\mathcal{A}_\eta$ with respect to $w'$ gives the analogous update from $w'$ to $w$:
\begin{equation*}
w = w' - \eta \nabla L(w').
\end{equation*}
There are only two circumstances where both criticality conditions are true simultaneously: (1) if $w$ and $w'$ are equal and are fixed points of gradient descent or (2) if $w$ and $w'$ are \textit{not} equal and form a two-period orbit. And notice that if we set $w = w'$, we have that:
\begin{equation}
\mathcal{A}_\eta(w,w) = 2 L(w).
\end{equation}

It is important to note the dependence of $\mathcal{A}_\eta$ on the step size. For a given loss function, the fixed points of gradient descent are independent of the step size. However, this is not true for the two-period orbits. If you change the step size, you change where the two-period orbits are.

\subsection{The Phase-Space Edge Coupling}
\label{sec:phase-coupling}

 We write heavy ball in phase-space form in the classical parametrization of \citet{polyak1964some}. Let $\beta \in [0, 1]$ denote the momentum parameter. The update equations are given as:
\begin{align}
	m_{t+1} &= \beta m_t - \nabla L(w_t), \label{eq:hb-m}\\
	w_{t+1} &= w_t + \eta m_{t+1}. \label{eq:hb-w}
\end{align}
where $w$ is the location in weight space and $m$ is the momentum. We write $T$ for the resulting map on phase space:
\begin{equation}
z_{t+1} = T(z_t)
\end{equation}
Fixed points of heavy-ball momentum satisfy the pair of equations:
\begin{align}
m &= 0, \\
\nabla L(w) &= 0.
\end{align}
The phase-space edge coupling $\mathcal{B}_{\eta, \beta}$ should be a functional that takes two phase-space points as arguments and whose critical points correspond to fixed points and two-point orbits.

Let $z = (w, m)$ and $z' = (w', m')$ denote two phase-space points. The edge coupling for heavy-ball momentum is given by:
\begin{equation}
	 \mathcal{B}_{\eta, \beta}(z, z') = L(w) + L(w') - \frac{1-\beta}{2\eta}\|w - w'\|^2 - \beta\,(m - m')\cdot(w - w') - \eta\beta\; m \cdot m' .
\label{eq:B-def}
\end{equation}
Note that when $z = z'$, we have that:
\begin{equation}
\mathcal{L}(w,m) \equiv \mathcal{B}_{\eta, \beta}(z,z) = 2 L(w) - \eta\beta \|m\|^2.
\label{eq:diagonal-H}
\end{equation}
Just as the weight-space edge coupling reduces to the loss function on its diagonal, the phase-space edge coupling reduces to a potential energy term minus a kinetic energy term. Up to an overall factor, this object is a \textit{Lagrangian}. Setting the derivatives of $\mathcal{L}$ with respect to position and momentum to zero gives:
\begin{align}
	\nabla_w \mathcal{L} &= 2 \nabla L(w) = 0, \\
	\nabla_m \mathcal{L} &= -2\eta\beta\, m = 0,
\end{align}
whose solutions are precisely the fixed points of the heavy-ball dynamics.

But we aren't just interested in the fixed points of heavy ball---we also want to understand the two-period orbits. Consider the general case where $z$ and $z'$ are not constrained to be equal. Taking the gradients of $\mathcal{B}_{\eta,\beta}$ with respect to $m$ and $w$:
\begin{align}
	\nabla_{m} \mathcal{B}_{\eta,\beta} &= -\beta\,\bigl(w - w' + \eta m'\bigr), \label{eq:grad-m}\\
	\nabla_{w} \mathcal{B}_{\eta,\beta} &= \nabla L(w) - \frac{1-\beta}{\eta}(w - w') - \beta\,(m - m'). \label{eq:grad-w}
\end{align}
For $\beta > 0$, by setting \cref{eq:grad-m} equal to zero, one can obtain:
$$w' = w + \eta m'$$
which is exactly the position update (\cref{eq:hb-w}) from $z$ to $z'$. And by substituting $w - w' = -\eta m'$ into \cref{eq:grad-w} and setting it equal to zero, one can obtain: 
$$m' = \beta m - \nabla L(w)$$
which is exactly the momentum update (\cref{eq:hb-m}). One can then see that:
\begin{equation}
	\nabla_{z} \mathcal{B}_{\eta,\beta}(z, z') = 0 \iff z' = T(z),
\label{eq:litman-property}
\end{equation}
And because $\mathcal{B}_{\eta, \beta}$ is symmetric under swapping $z$ and $z'$, we also have that:
\begin{equation}
\nabla_{z'} \mathcal{B}_{\eta,\beta}(z, z') = 0 \iff z = T(z')
\end{equation}
As was the case with gradient descent, there are only two circumstances where every critical condition of $\mathcal{B}_{\eta, \beta}$ is satisfied: $z = z'$ is a fixed point, or $z \neq z'$ and $(z, z')$ is a two-period orbit.

\subsection{Reduction to Position Space}
\label{sec:momentum-reduction}

We can recover a position-space edge coupling from the phase-space edge coupling by eliminating the momentum coordinates. To do so, we can use the momentum criticality conditions
\begin{align}
m &= \frac{1}{\eta} (w-w') \label{eq:momentum-criticality}\\
m' &= \frac{1}{\eta} (w'-w)
\end{align}
to express $m$ and $m'$ in terms of the positions, and then substitute back into $\mathcal{B}_{\eta, \beta}$. We then obtain:
\begin{equation}
\mathcal{A}_{\eta,\beta}(w, w') \equiv	\operatorname*{stat}_{m, m'} \mathcal{B}_{\eta,\beta}(z, z') = L(w) + L(w') - \frac{1+\beta}{2\eta}\|w-w'\|^2 
\label{eq:A-reduced}
\end{equation}
Critical points of the position-space edge coupling correspond to points in weight space where there exists \textit{some} corresponding pair of momenta $(m,m')$ such that $(w,w')$ form a two-period orbit. And by inspection, one can see that we recover Litman's edge coupling for gradient descent when $\beta = 0$.

\subsection{Nesterov Momentum}
\label{sec:nesterov}

Nesterov momentum \citep{nesterov1983method} is an alternative to heavy ball momentum. It differs in that the
gradient is evaluated at a look-ahead point, displaced from the current iterate along the momentum \citep{sutskever2013importance}:
\begin{align}
	m_{t+1} &= \beta m_t - \nabla L(w_t + \eta\beta m_t), \label{eq:nag-m}\\
	w_{t+1} &= w_t + \eta m_{t+1}. \label{eq:nag-w}
\end{align}
Nesterov momentum also admits a phase-space coupling:
\begin{equation}
	\mathcal{B}^{\mathrm{NAG}}_{\eta,\beta}(z,z') = L(w+\eta\beta m) + L(w'+\eta\beta m')
	- \frac{1-\beta}{2\eta}\,\|w-w'\|^2
	- \beta\,(m-m')\cdot(w-w')
	- \frac{\eta\beta^2}{2}\big(\|m\|^2+\|m'\|^2\big).
	\label{eq:nag-B}
\end{equation}
For simplicity, we work with heavy ball momentum throughout this paper. The analogous results for Nesterov momentum are collected in \cref{app:nesterov}.

\section{Coordinate Transformations}
\label{sec:coordinates}

For any function, its critical points are invariant under changes of coordinates that are smooth and invertible. We then have that the edge coupling can be expressed in different coordinates while preserving its key property.

\subsection{Invariance of Critical Points}
\label{sec:invariance}

Let $U$ denote a smooth invertible change of coordinates on the doubled phase space. Concretely, $U$ takes $(z, z')$ to a point $u = U(z, z')$ of a $4d$-dimensional space with inverse $U^{-1}(u) = (z, z')$. The coupling in the new coordinates is given as 
\begin{equation}
\mathcal{B}_U \coloneqq \mathcal{B}_{\eta,\beta} \circ U^{-1}
\end{equation}
We can compute the gradient of $\mathcal{B}_{U}$ by using the chain rule:
\begin{equation}
	\nabla_u \mathcal{B}_U(u) = \bigl[D U^{-1}(u)\bigr]^{\top}\, \nabla_{(z,z')} \mathcal{B}_{\eta,\beta}\bigl(U^{-1}(u)\bigr).
	\label{eq:chain-rule}
\end{equation}
Because $D U^{-1}$ is invertible, it follows that:
\begin{equation*}
	\nabla_u \mathcal{B}_U = 0 \iff \nabla_{(z,z')} \mathcal{B}_{\eta,\beta} = 0,
\end{equation*}
so critical points correspond bijectively across the change of coordinates.

\subsection{Centered Coordinates}
\label{sec:centered}

One natural change of coordinates is to centered variables:
\begin{align}
	\bar{z} &= \begin{pmatrix} \bar{w} \\ \bar{m} \end{pmatrix} = \frac{z + z'}{2}, \\
	\Delta &= \begin{pmatrix} \delta \\ \gamma \end{pmatrix} = \frac{z - z'}{2}.
\end{align}
where $\bar{z}$ is the midpoint of the phase-space variables and $\Delta$ is the half-displacement between them. In centered coordinates, our phase-space coupling becomes:
\begin{equation}
\mathcal{B}_{\eta,\beta}(\bar z + \Delta,\, \bar z - \Delta) 
= L(\bar w + \delta) + L(\bar w - \delta) - \frac{2(1-\beta)}{\eta}\|\delta\|^2 - 4\beta\,\gamma \cdot \delta - \eta\beta\left(\|\bar{m}\|^2 - \|\gamma\|^2\right)
\end{equation}
It is instructive to take derivatives of $\mathcal{B}$ with respect to the centered variables to obtain criticality conditions. Taking the partial derivative with respect to $\bar{m}$, we have:
\begin{equation*}
	\frac{\partial \mathcal{B}}{\partial \bar{m}} = -2\eta\beta\, \bar{m} = 0.
\end{equation*}
For $\beta > 0$, criticality forces $\bar m = 0$: every two-point orbit carries zero mean-momentum.

Taking the derivative with respect to $\gamma$:
\begin{equation*}
	\frac{\partial \mathcal{B}}{\partial \gamma} = -4\beta\,\delta + 2\eta\beta\,\gamma = 0,
\end{equation*}
from which one can deduce that
\begin{equation*}
	\gamma = \frac{2}{\eta}\,\delta.
\end{equation*}
This is equivalent to the momentum criticality condition $\eta m' = w' - w$.

If we perform the reduction to position space, we recover Litman's edge coupling in centered coordinates:
\begin{equation}
	\Psi(\bar{w}, \delta) \coloneqq \mathcal{A}_{\eta,\beta}(\bar w - \delta,\, \bar w + \delta) = L(\bar{w} - \delta) + L(\bar{w} + \delta) - \frac{2(1+\beta)}{\eta}\|\delta\|^2.
	\label{eq:Psi-def}
\end{equation}
And if we take derivatives of $\Psi$, we obtain the criticality conditions:
\begin{align}
	\nabla_{\bar w}\Psi &= \nabla L(\bar w - \delta) + \nabla L(\bar w + \delta) =0 \\
	\nabla_{\delta}\Psi &= \nabla L(\bar w + \delta) - \nabla L(\bar w - \delta) - \frac{4(1+\beta)}{\eta}\,\delta = 0.
\end{align}
The $\bar{w}$ criticality condition encodes the requirement that, for two-point orbits, the \textit{average} gradient vanishes. The $\delta$ criticality condition is essentially a self-consistency equation: the difference of the gradients must match the half-displacement between the two points.

\section{Stability and the Hessian}
\label{sec:hessian}

The phase-space edge coupling not only encodes the \textit{existence} of two-point orbits via its critical points---it also encodes their linear stability via its Hessian. 

\subsection{Linearization of the Two-Step Map}
\label{sec:linearization}

Consider the map corresponding to the two-step update:
\begin{equation*}
T_2(z_t) \equiv [T \circ T ](z_t)= z_{t+2}
\end{equation*}
The fixed points of $T_2$ exactly correspond to the fixed points and two-point orbits of heavy ball. 

Let $H$ denote the Hessian of the loss function, and let $J$ denote the Jacobian of the one-step update. $J$ can be expressed as a function of the Hessian evaluated at that point:
\begin{equation}
J(H) \coloneqq DT(z) = \begin{pmatrix} I_d - \eta H & \eta\beta I_d \\ -H & \beta I_d \end{pmatrix}.
\label{eq:one-step-jacobian}
\end{equation}
Let $M$ denote the Jacobian of the two-step update. Consider a pair of phase-space
points $(z, z')$ forming a two-period orbit. By the chain rule, $M$ evaluated at
$z$ is the product of the one-step Jacobians evaluated at both points:
\begin{equation}
	M(z) \coloneqq D T_2(z) = J(H_{w'})\, J(H_w).
	\label{eq:monodromy}
\end{equation}
Let $\{\lambda_i\}$ denote the eigenvalues of $M$. The two-period orbit is linearly stable when every eigenvalue $\lambda_i$ lies inside the unit circle. It is important to note that while $M(z)$ and $M(z')$ differ whenever the one-step Jacobians at $z$ and $z'$ fail to commute, the two matrices are conjugate and hence share the same spectrum.

Even though $M$ acts on phase space, the momentum variables can be eliminated to
yield an eigenvalue equation defined in terms of position-space operators alone:
\begin{equation}
	\det\!\left((\lambda + \beta)^2 I_d - \lambda A_{w'} A_{w}\right) = 0,
\end{equation}
where
\begin{equation}
A_w = (1+\beta)I_d - \eta H_w.
\end{equation}
Unlike ordinary eigenvalue problems, which involve a linear pencil (a parametric
family of matrices depending affinely on $\lambda$), this eigenvalue problem involves a \textit{quadratic}
pencil:
\begin{equation}
	W(\lambda) = (\lambda + \beta)^2 I_d - \lambda A_{w'} A_{w}
\end{equation}
The phase-space and position-space eigenvalue problems are equivalent. For all values of $\lambda$, the determinant of the linear pencil defined on the space of  phase-space operators equals the determinant of the quadratic pencil defined on the space of position-space operators.
\begin{equation}
\det (M - \lambda I_{2d}) = \det \left((\lambda + \beta)^2 I_d - \lambda A_{w'} A_{w}\right)
\end{equation}
 More details are provided in \cref{app:pencil-proof}.

\subsection{The Hessian of the Edge Coupling}
\label{sec:hessian-B}

We want to derive a condition for the Hessian of $\mathcal{B}_{\eta,\beta}$ that corresponds to linear stability. It is helpful to compute the Hessian in block form corresponding to the coordinates $(w, w', m, m')$. For the second-order partial derivatives, we have:
\begin{align*}
\frac{\partial^2 \mathcal{B}}{\partial w^2} &= H_w - \frac{1-\beta}{\eta}\, I_d, &
\frac{\partial^2 \mathcal{B}}{\partial w\, \partial w'} &= \frac{1-\beta}{\eta}\, I_d, \\
\frac{\partial^2 \mathcal{B}}{\partial w\, \partial m} &= -\beta\, I_d, &
\frac{\partial^2 \mathcal{B}}{\partial w\, \partial m'} &= \beta\, I_d, \\
\frac{\partial^2 \mathcal{B}}{\partial m^2} &= 0, &
\frac{\partial^2 \mathcal{B}}{\partial m\, \partial m'} &= -\eta\beta\, I_d.
\end{align*}
Expressions for the other second-order partial derivatives can be found by using the $(z \leftrightarrow z')$ symmetry of $\mathcal{B}$. Assembling the blocks, we then have:
\begin{equation}
	\nabla^2 \mathcal{B}_{\eta, \beta} =
	\begin{pmatrix}
		H_w - \frac{1-\beta}{\eta} I & \frac{1-\beta}{\eta} I & -\beta I & \beta I \\[3pt]
		\frac{1-\beta}{\eta} I & H_{w'} - \frac{1-\beta}{\eta} I & \beta I & -\beta I \\[3pt]
		-\beta I & \beta I & 0 & -\eta\beta I \\[3pt]
		\beta I & -\beta I & -\eta\beta I & 0
	\end{pmatrix}
	\label{eq:hessian-B}
\end{equation}
in the block order $(w, w',m, m')$.

Analogous to how we derived a momentum-reduced coupling
$\mathcal{A}_{\eta,\beta}$, we would like to obtain a reduced expression for $\nabla^2 \mathcal{B}_{\eta, \beta}$. To do so, we use the Schur complement.
Grouping the coordinates into positions $(w, w')$ and momenta $(m, m')$, the $\nabla^2 \mathcal{B}_{\eta, \beta}$ can be expressed in $2 \times 2$ block form:
\begin{equation}
	\nabla^2 \mathcal{B} =
	\begin{pmatrix}
		\partial^2_{(w,w')} \mathcal{B} & \partial_{(w,w')} \partial_{(m,m')} \mathcal{B} \\[3pt]
		\partial_{(m,m')} \partial_{(w,w')} \mathcal{B} & \partial^2_{(m,m')} \mathcal{B}
	\end{pmatrix}.
	\label{eq:hessian-block-form}
\end{equation}
Let $S$ denote the Schur complement. Since $\eta\beta \neq 0$, the momentum block is invertible---so the Schur
complement of $\partial^2_{(m,m')}\mathcal{B}$ in $\nabla^2\mathcal{B}$
is well defined:
\begin{equation}
	S = \partial^2_{(w,w')} \mathcal{B}
	\;-\;
	\partial_{(w,w')} \partial_{(m,m')} \mathcal{B}
	\,\bigl(\partial^2_{(m,m')} \mathcal{B}\bigr)^{-1}
	\partial_{(m,m')} \partial_{(w,w')} \mathcal{B}.
	\label{eq:schur-def}
\end{equation}
Substituting the blocks into our expression for $S$, we
obtain:
\begin{equation}
	S =
	\begin{pmatrix}
		H_w - \frac{1+\beta}{\eta} I & \frac{1+\beta}{\eta} I \\[3pt]
		\frac{1+\beta}{\eta} I & H_{w'} - \frac{1+\beta}{\eta} I
	\end{pmatrix}.
	\label{eq:schur-explicit}
\end{equation}
One should note that $S$ is \textit{precisely} the Hessian of $\mathcal{A}_{\eta,\beta}$.
\begin{equation}
	\nabla^2 \mathcal{A}_{\eta, \beta} = S
\end{equation}
The Schur complement induces the block LDL factorization of the Hessian of $\mathcal{B}_{\eta, \beta}$:
\begin{equation}
	\nabla^2 \mathcal{B} =
	\begin{pmatrix}
		I & \partial_{(w,w')} \partial_{(m,m')} \mathcal{B}
		\,\bigl(\partial^2_{(m,m')} \mathcal{B}\bigr)^{-1} \\[3pt]
		0 & I
	\end{pmatrix}
	\begin{pmatrix}
		S & 0 \\[3pt]
		0 & \partial^2_{(m,m')} \mathcal{B}
	\end{pmatrix}
	\begin{pmatrix}
		I & 0 \\[3pt]
		\bigl(\partial^2_{(m,m')} \mathcal{B}\bigr)^{-1}
		\partial_{(m,m')} \partial_{(w,w')} \mathcal{B} & I
	\end{pmatrix}.
	\label{eq:ldl}
\end{equation}
Because the outer factors are unit triangular, \cref{eq:ldl} is a \textit{congruence transformation}---we have that the determinant and the inertia of $\nabla^2\mathcal{B}$ match the corresponding quantities for the block-diagonal middle factor.

\subsection{Determinant}
\label{sec:determinant}

We would like to understand how $\nabla^2 \mathcal{B}$ encodes linear stability. To do so, we will obtain an expression for the determinant of $\mathcal{B}$ in terms of $M$, the two-step Jacobian.

There is a useful identity for computing the determinant of block matrices. Let $X$ be a matrix of the form
\begin{equation}
	X = \begin{pmatrix} A & B \\ C & D \end{pmatrix},
\end{equation}
where $A, B, C, D$ are square blocks of the same size. If $C$ and $D$ commute, we then have that:
\begin{equation}
\det X = \det(AD - BC)
\end{equation}
Let $\kappa = \tfrac{1+\beta}{\eta}$. For the determinant of $S$, we have:
\begin{equation}
\det S = \det[H_{w}H_{w'} - \kappa (H_w + H_{w'})]
\end{equation}
Recall the quadratic pencil $W(\lambda)$, the parametric family of matrices which featured in our position-space eigenvalue equation for $M$. When $\lambda=1$, we have that the determinant of that matrix is equal to
\begin{equation}
\det W(1) = \eta^{2d} \det \left [ \kappa (H_w + H_{w'}) - H_{w'} H_{w} \right]
\end{equation}
As the determinant of our quadratic pencil matches that of the linear pencil for the corresponding value of $\lambda$, it then follows that we have that:
\begin{equation}
\det (M-I) = \eta^{2d} \det \left [ \kappa (H_w + H_{w'}) - H_{w'} H_{w} \right]
\end{equation}
Using the fact that $\det (XY - Z) = \det (YX - Z)$ for symmetric $X$, $Y$, $Z$---the two matrices are transposes of one another---we can then deduce:
\begin{equation}
	\det S =  \frac{(-1)^d}{\eta^{2d}} \det (M-I) 
\end{equation}
Using the LDL factorization, we can then compute the determinant of $\nabla^2 \mathcal{B}$:
\begin{equation}
\det \nabla^2 \mathcal{B} = \det S \cdot \det \partial^2_{(m,m')} \mathcal{B} = \beta^{2d} \det (M - I)
\end{equation}
Recall that linear stability corresponds to when the eigenvalues of $M$ are contained within the unit circle. The determinant of $\nabla^2 \mathcal{B}$ undergoes a sign flip \textit{precisely} when an eigenvalue of $M$ crosses $1$, so $\mathcal{B}$ encodes the period-doubling bifurcation from a stable fixed point to a stable two-period orbit. It cannot, however, detect the subsequent transitions from the two-period orbit to higher-period orbits, since these
occur when an eigenvalue of $M$ exits the unit circle at complex or negative real values.

\subsection{Inertia}
\label{sec:inertia-stability}
The inertia of a symmetric matrix is the triple
\begin{equation}
\In(X) = (n_+, n_-, n_0)	
\end{equation}
counting its positive, negative, and zero eigenvalues. By Sylvester's law of inertia, the inertia is invariant under congruence transformations $X \mapsto Y^\top X Y$ for invertible $Y$. Applied to the LDL factorization, this identifies the inertia of
$\nabla^2\mathcal{B}$ with that of the block-diagonal middle factor:
\begin{equation}
	\In\bigl(\nabla^2\mathcal{B}\bigr)
	= \In(S) + \In\bigl(\partial^2_{(m,m')}\mathcal{B}\bigr).
\end{equation}
The inertia is also invariant under the change of coordinates $U$. At
a critical point, the chain rule gives the congruence
\begin{equation}
	\nabla^2 \mathcal{B}_U = [ D U^{-1}]^\top \bigl(\nabla^2 \mathcal{B}_{\eta,\beta}\bigr) [D U^{-1}],
\end{equation}
so the two Hessians have the same inertia, and their determinants
agree up to the positive factor $(\det D U^{-1})^2$. Both the inertia and the sign of the determinant are therefore intrinsic to the orbit,
independent of the choice of coordinates.

Note that the momentum block is the fixed matrix 
$$-\eta\beta \begin{pmatrix} 0 & I \\ I & 0 \end{pmatrix}$$
with eigenvalues $\pm\eta\beta$, each of multiplicity $d$. Assuming that
$\eta\beta \neq 0$, its inertia is $(d, d, 0)$---irrespective of
the orbit. All orbit-dependent inertia is carried by $S$.

We will now show how the inertia encodes linear stability. We claim that every linearly stable two-point orbit is a critical point of $\mathcal{B}_{\eta,\beta}$ at which $S$ has inertia $(d, d, 0)$---a
perfectly balanced saddle.

To see why, extend the Schur complement (\cref{eq:schur-explicit}) to the
one-parameter family
\begin{equation}
	S(\lambda) \coloneqq
	\begin{pmatrix}
		H_w - \frac{1+\beta}{\eta} I & \nu_\lambda I \\[3pt]
		\nu_\lambda I & H_{w'} - \frac{1+\beta}{\eta} I
	\end{pmatrix},
	\qquad
	\nu_\lambda \coloneqq \frac{\lambda + \beta}{\eta \sqrt{\lambda}},
	\qquad \lambda > 0,
\end{equation}
which recovers the Schur complement at $\lambda = 1$. One can relate the determinant of $S(\lambda)$ to the determinant of the linear pencil defined on the space of phase-space operators:
\begin{equation}
	\det\bigl(M - \lambda I_{2d}\bigr)
	= (-1)^d \lambda^d \eta^{2d} \det S(\lambda).
	\label{eq:pencil-det}
\end{equation}
Now suppose the orbit is linearly stable so that $M$ has no eigenvalue
on or outside the unit circle---in particular, no real eigenvalue greater than 1. We then have that $S(\lambda)$ is nondegenerate for
\textit{all} $\lambda \in [1, \infty)$. Since the eigenvalues of $S(\lambda)$ vary
continuously in $\lambda$ and none of them crosses zero, the inertia of $S(\lambda)$ is constant on $[1,\infty)$. 

But as $\lambda \to \infty$, the off-diagonal
coupling $\nu_\lambda \to \infty$ dominates the $\lambda$-independent diagonal blocks. We have that the eigenvalues of $S(\lambda)$ approach $\pm\nu_\lambda$, each with multiplicity $d$---so the inertia of $S(\infty)$ is $(d,d,0)$. By continuity, we have that $S =
S(1)$ has inertia $(d,d,0)$. And by the factorization, it follows that $\nabla^2\mathcal{B}$ has inertia $(2d, 2d, 0)$.

\section{The St\"ormer--Verlet Action}
\label{sec:mechanics}

\citet{litman2026origin} remarked that the edge coupling can be viewed as ``a discrete generating function encoding gradient descent in boundary form''. We make this explicit by demonstrating that the edge coupling is connected to the symmetric Verlet action.

\subsection{Recap of Discrete Mechanics}
\label{sec:dm-recap}

We will briefly recap discrete mechanics.

Given a potential function $L$, Newton's second law states that the acceleration multiplied by the mass of the particle equals the negative gradient of the potential:
\begin{equation}
	\mu \ddot{w} = -\nabla L,
\end{equation}
where $w$ is the position and $\mu$ is the mass.

One can then generate the trajectory of the particle by specifying two initial conditions (usually the initial position $w_0$ and initial velocity $v_0$) and integrating the ODE. To implement this on a computer, one must choose a discrete step size $\Delta t$ and update the position and velocity accordingly. One canonical choice is Euler discretization:
\begin{align}
	v_{t+\Delta t} &= v_t + \Delta t \left( -\frac{1}{\mu} \nabla L(w_t) \right), \\
	w_{t+\Delta t} &= w_t + \Delta t \, v_t .
\end{align}
However, there is an alternative approach. Classical mechanics admits a variational formulation---\textit{the principle of least action}. It dictates that a particle follows the trajectory that extremizes the action, a functional on the space of paths:
\begin{equation}
	\mathcal{S}[w(t)] = \int \mathcal{L}(w, v, t) \, dt .
\end{equation}
The integrand $\mathcal{L}$ is called the \textit{Lagrangian}. For a unit-mass particle, we have that:
\begin{equation*}
	\mathcal{L}(w, v) = \tfrac12 \|v\|^2 - L(w).
\end{equation*}
It is the difference between the kinetic and potential energy.

An alternative approach to simulating the dynamics of the particle is to discretize the \textit{action} in time, rather than the ODE. There are many ways one could discretize the action integral. One choice is to approximate the velocity over a single step by the finite difference $(w_{n+1} - w_n)/\Delta t$ and the potential term by the trapezoidal rule. This choice gives us the following \emph{discrete Lagrangian}:
\begin{equation}
	L_d(w_n, w_{n+1}) = \frac{1}{2\Delta t}\|w_{n+1} - w_n\|^2 - \frac{\Delta t}{2}\bigl(L(w_n) + L(w_{n+1})\bigr).
	\label{eq:Ld}
\end{equation}
Visualizing the action as an integral, the discrete Lagrangian gives the little block of area corresponding to one step $\Delta t$. The action is now a function of the $N-1$ interior points, with the endpoints $w_0$ and $w_N$ held fixed:
\begin{equation}
	\mathcal{S}_d(w_1, \dots, w_{N-1}) = \sum_{n=1}^{N} L_{d}(w_{n-1}, w_n).
\end{equation}
The discrete Hamilton's principle requires $\mathcal{S}_d$ to be stationary with respect to perturbations of every interior point. Since $w_n$ appears in only two terms of the sum, this gives for $n = 1, \dots, N-1$:
\begin{equation}
	\frac{\partial}{\partial w_n}\bigl[L_d(w_{n-1}, w_n) + L_d(w_n, w_{n+1})\bigr] = 0.
\end{equation}
Computing the two derivatives,
\begin{align*}
	\frac{\partial}{\partial w_n} L_d(w_{n-1}, w_n) &= \frac{1}{\Delta t}(w_n - w_{n-1}) - \frac{\Delta t}{2}\nabla L(w_n), \\
	\frac{\partial}{\partial w_n} L_d(w_n, w_{n+1}) &= -\frac{1}{\Delta t}(w_{n+1} - w_n) - \frac{\Delta t}{2}\nabla L(w_n),
\end{align*}
and substituting into the stationarity condition above yields the St\"ormer--Verlet scheme \citep{verlet1967computer, hairer2003acta}, also known as the leapfrog integrator:
\begin{equation}
	w_{n+1} - 2w_n + w_{n-1} = -(\Delta t)^2\, \nabla L(w_n).
	\label{eq:verlet}
\end{equation}
Setting $\Delta t = \sqrt{\eta}$ recovers heavy ball with $\beta = 1$.

\subsection{Edge Coupling as Discrete Action}
\label{sec:period2-action}

Consider the scenario in which the leapfrog integrator is trapped in a period-two orbit, alternating between two points $w$ and $w'$. The total action of the trajectory then splits into identical two-step blocks,
\begin{equation}
	\mathcal{S}_d = L_d(w, w') + L_d(w', w) + L_d(w, w') + \cdots = \frac{N}{2}\, \mathcal{S}_d^{(2)}(w, w'),
\end{equation}
where we define the \emph{two-period action}
\begin{equation}
	\mathcal{S}_d^{(2)}(w, w') \coloneqq L_d(w, w') + L_d(w', w) = \frac{1}{\Delta t}\|w - w'\|^2 - \Delta t\,\bigl(L(w) + L(w')\bigr).
	\label{eq:Sd2}
\end{equation}
Discrete Hamilton's principle requires stationarity with respect to every interior point. The even-indexed points impose $\partial_w \mathcal{S}_d^{(2)} = 0$, and the odd-indexed points impose $\partial_{w'} \mathcal{S}_d^{(2)} = 0$. Therefore, stationarity of the whole trajectory corresponds to stationarity of the two-period action $\mathcal{S}_d^{(2)}$ with respect to both of its arguments.

We can express the position-space edge coupling in terms of the two-period action:
\begin{align}
	\mathcal{A}_{\eta, \beta} &= L(w) + L(w') - \frac{1+\beta}{2\eta} \|w - w'\|^2 \nonumber \\
	&= -\sqrt{\frac{1+\beta}{2\eta}}\left[\sqrt{\frac{1+\beta}{2\eta}}\,\|w - w'\|^2 - \sqrt{\frac{2\eta}{1+\beta}}\,\bigl(L(w) + L(w')\bigr)\right] \nonumber\\
	&= -\frac{1}{\Delta t}\;\mathcal{S}_d^{(2)}(w, w') \bigg|_{\Delta t = \sqrt{2\eta/(1+\beta)}}.
	\label{eq:position-action-identity}
\end{align}
Because the edge coupling and the two-period action coincide up to an overall factor, they have the same critical points. We then have that the two-period orbits of heavy ball with stepsize $\eta$ and momentum coefficient $\beta$ are \textit{exactly} the two-period orbits of the leapfrog integrator at $\Delta t = \sqrt{2\eta/(1+\beta)}$.

\subsection{The Phase-Space Action}
\label{sec:phase-space-action}

We can derive a similar identity for the phase-space edge coupling. Consider the Hamiltonian
\begin{equation}
	\mathcal{H}(w, m) = \tfrac{1}{2} \|m\|^2 + L(w),
\end{equation}
which represents the total energy of the system. The Lagrangian and the Hamiltonian are connected by the Legendre transform, in which one trades the velocity $v$ for the momentum $m$:
\begin{equation}
	\mathcal{L}_{\mathrm{ph}}(w, m) = m \cdot v - \mathcal{H}(w, m), \qquad m = \frac{\partial \mathcal{L}}{\partial v} = v .
\end{equation}
Substituting this into the action yields the \textit{Hamilton--Pontryagin action}:
\begin{equation}
	\mathcal{S}_{\mathrm{ph}}[w(t), m(t)] = \int \bigl( m \cdot \dot{w} - \mathcal{H}(w, m) \bigr) \, dt ,
\end{equation}
where $w(t)$ and $m(t)$ are varied independently.

We can also derive the leapfrog integrator by discretizing this phase-space action in time. Placing the momentum at the midpoints of the steps and approximating $\dot{w}$ over one step by $(w_n - w_{n-1})/\Delta t$ gives the \emph{discrete phase-space Lagrangian}:
\begin{equation}
	\mathcal{L}_{\mathrm{ph}, d}(w_{n-1}, w_n, m_{n-1/2}) = m_{n-1/2} \cdot (w_n - w_{n-1}) - \Delta t \left( \tfrac{1}{2}\|m_{n-1/2}\|^2 + \tfrac{1}{2}\bigl(L(w_{n-1}) + L(w_n)\bigr) \right).
\end{equation}
The discrete action is given by the sum over steps:
\begin{equation}
	\mathcal{S}_{\mathrm{ph}, d} = \sum_{n=1}^{N} \mathcal{L}_{\mathrm{ph}, d}(w_{n-1}, w_n, m_{n-1/2}).
\end{equation}
Hamilton's principle requires that the action be stationary with respect to perturbations of every $m_{n-1/2}$ and every interior $w_n$, keeping the endpoints $w_0$ and $w_N$ fixed. Stationarity with respect to $m_{n-1/2}$ gives
\begin{equation}
m_{n-1/2} = \frac{w_n - w_{n-1}}{\Delta t}.
\end{equation}
So the momentum, though introduced as an independent variable, is constrained to be equal to the velocity. Stationarity with respect to $w_n$ gives
\begin{equation}
m_{n+1/2} = m_{n-1/2} - \Delta t\, \nabla L(w_n).
\end{equation}
As expected, the momentum update is in the direction of the force.

Consider the scenario in which the leapfrog integrator is locked in a period-two orbit, alternating between $w$ and $w'$. The two-period action is obtained by summing two consecutive discrete Lagrangians:
\begin{align}
	\mathcal{S}^{(2)}_{\mathrm{ph}, d}(w, m, w', m') &= \mathcal{L}_{\mathrm{ph}, d}(w, w', m') + \mathcal{L}_{\mathrm{ph}, d}(w', w, m) \\
	&= (m - m') \cdot (w - w') - \Delta t \left[ \tfrac12\|m\|^2 + \tfrac12\|m'\|^2 + L(w) + L(w') \right].
	\label{eq:phase-action}
\end{align}
We can express the phase-space edge coupling in terms of $\mathcal{S}^{(2)}_{\mathrm{ph}, d}$:
\begin{align}
	\mathcal{B}_{\eta, \beta}(z, z') &= L(w) + L(w') - \frac{1-\beta}{2\eta}\|w - w'\|^2 - \beta\,(m - m')\cdot(w - w') - \eta\beta\; m \cdot m' \nonumber\\
	&= -\frac{1}{\Delta t}\, \mathcal{S}^{(2)}_{\mathrm{ph}, d}\bigl(w, \beta \Delta t\, m,\; w', \beta \Delta t\, m'\bigr) \nonumber \\
	&\qquad -\; \frac{\eta\beta(1 + 3\beta)}{4(1+\beta)}\,\|m + m'\|^2 \;+\; \frac{\eta\beta(1 - \beta)}{4(1+\beta)}\,\|m - m'\|^2 \;-\; \frac{1-\beta}{2\eta}\,\|w - w'\|^2 .
	\label{eq:gauge-residual}
\end{align}
One can see that there are residual terms. That's because the two functionals only share the same critical \emph{positions}. Their critical momenta differ by a fixed rescaling. More details are provided in \cref{app:action-proof}.

\section{Conclusion}
\label{sec:conclusion}

We have presented a variational framework for understanding momentum algorithms at the edge of stability. We hope that this sheds light on the near-two-periodic behavior observed in this regime.

\paragraph{Future work.} Natural next steps are a variational account of higher-period orbits.

\paragraph{Acknowledgments}

We used various iterations of Claude---Opus 4.8, Opus 5.0, and Fable 5.0---in the process of brainstorming, drafting, and editing this paper.

\bibliography{symplectic-structure-at-the-edge-of-stability}

@article{litman2026origin,
  title={The Origin of Edge of Stability},
  author={Litman, Elon},
  journal={arXiv preprint arXiv:2604.20446},
  year={2026}
}

@article{regis2026rod,
  title={Rod flow: a continuous-time model for gradient descent at the edge of stability},
  author={Regis, Eric and Chewi, Sinho},
  journal={arXiv preprint arXiv:2602.01480},
  year={2026}
}

@article{regis2026adamrod,
  title={A Rod Flow Model for Adam at the Edge of Stability},
  author={Regis, Eric and Chewi, Sinho},
  journal={arXiv preprint arXiv:2605.06821},
  year={2026}
}

@inproceedings{chen2023beyond,
  title={Beyond the Edge of Stability via Two-step Gradient Updates},
  author={Chen, Lei and Bruna, Joan},
  booktitle={International Conference on Machine Learning},
  volume={202},
  pages={4330--4391},
  year={2023}
}

@inproceedings{ahn2022understanding,
  title={Understanding the Unstable Convergence of Gradient Descent},
  author={Ahn, Kwangjun and Zhang, Jingzhao and Sra, Suvrit},
  booktitle={International Conference on Machine Learning},
  year={2022}
}

@inproceedings{arora2022understanding,
  title={Understanding Gradient Descent on the Edge of Stability in Deep Learning},
  author={Arora, Sanjeev and Li, Zhiyuan and Panigrahi, Abhishek},
  booktitle={International Conference on Machine Learning},
  year={2022}
}

@inproceedings{damian2023selfstabilization,
  title={Self-Stabilization: The Implicit Bias of Gradient Descent at the Edge of Stability},
  author={Damian, Alex and Nichani, Eshaan and Lee, Jason D.},
  booktitle={International Conference on Learning Representations},
  year={2023}
}

@inproceedings{cohen2025central,
  title={Understanding Optimization in Deep Learning with Central Flows},
  author={Cohen, Jeremy M. and Damian, Alex and Talwalkar, Ameet and Kolter, J. Zico and Lee, Jason D.},
  booktitle={International Conference on Learning Representations},
  year={2025}
}

@article{andreyev2026momentum,
  title={Momentum Further Constrains Sharpness at the Edge of Stochastic Stability},
  author={Andreyev, Arseniy and Ananthkumar, Advikar and Walden, Marc and Poggio, Tomaso and Beneventano, Pierfrancesco},
  journal={arXiv preprint arXiv:2604.14108},
  year={2026}
}

@article{phunyaphibarn2024gradient,
  title={Gradient descent with {P}olyak's momentum finds flatter minima via large catapults},
  author={Phunyaphibarn, Prin and Lee, Junghyun and Wang, Bohan and Zhang, Huishuai and Yun, Chulhee},
  journal={arXiv preprint arXiv:2311.15051},
  year={2024}
}

@article{wibisono2016variational,
  title={A Variational Perspective on Accelerated Methods in Optimization},
  author={Wibisono, Andre and Wilson, Ashia C. and Jordan, Michael I.},
  journal={Proceedings of the National Academy of Sciences},
  volume={113},
  number={47},
  pages={E7351--E7358},
  year={2016}
}

@inproceedings{franca2020conformal,
  title={Conformal Symplectic and Relativistic Optimization},
  author={Fran{\c{c}}a, Guilherme and Sulam, Jeremias and Robinson, Daniel P. and Vidal, Ren{\'e}},
  booktitle={Advances in Neural Information Processing Systems},
  year={2020}
}

@article{muehlebach2021optimization,
  title={Optimization with Momentum: Dynamical, Control-Theoretic, and Symplectic Perspectives},
  author={Muehlebach, Michael and Jordan, Michael I.},
  journal={Journal of Machine Learning Research},
  volume={22},
  number={73},
  pages={1--50},
  year={2021}
}

@article{marion2026edge,
  title={Edge Flow: A Tractable and Predictive Continuous-Time Model for Gradient Descent at the Edge of Stability},
  author={Marion, Pierre},
  journal={arXiv preprint arXiv:2606.18080},
  year={2026}
}

@inproceedings{cohen2021gradient,
  title={Gradient descent on neural networks typically occurs at the edge of stability},
  author={Cohen, Jeremy and Kaur, Simran and Li, Yuanzhi and Kolter, J. Zico and Talwalkar, Ameet},
  booktitle={International Conference on Learning Representations},
  year={2021}
}

@article{cohen2022adaptive,
  title={Adaptive gradient methods at the edge of stability},
  author={Cohen, Jeremy M. and Ghorbani, Behrooz and Krishnan, Shankar and Agarwal, Naman and Medapati, Sourabh and Badura, Michal and Suo, Daniel and Cardoze, David and Nado, Zachary and Dahl, George E. and Gilmer, Justin},
  journal={arXiv preprint arXiv:2207.14484},
  year={2022}
}

@inproceedings{jastrzebski2020breakeven,
  title={The Break-Even Point on Optimization Trajectories of Deep Neural Networks},
  author={Jastrz{\k{e}}bski, Stanis{\l}aw and Szymczak, Maciej and Fort, Stanislav and Arpit, Devansh and Tabor, Jacek and Cho, Kyunghyun and Geras, Krzysztof},
  booktitle={International Conference on Learning Representations},
  year={2020}
}

@article{polyak1964some,
  title={Some methods of speeding up the convergence of iteration methods},
  author={Polyak, B. T.},
  journal={USSR Computational Mathematics and Mathematical Physics},
  volume={4},
  number={5},
  pages={1--17},
  year={1964}
}

@article{marsden2001discrete,
  title={Discrete mechanics and variational integrators},
  author={Marsden, Jerrold E. and West, Matthew},
  journal={Acta Numerica},
  volume={10},
  pages={357--514},
  year={2001}
}

@book{hairer2006geometric,
  title={Geometric numerical integration: structure-preserving algorithms for ordinary differential equations},
  author={Hairer, Ernst and Lubich, Christian and Wanner, Gerhard},
  year={2006},
  publisher={Springer}
}

@article{hairer2003acta,
  title={Geometric numerical integration illustrated by the {S}t{\"o}rmer--{V}erlet method},
  author={Hairer, Ernst and Lubich, Christian and Wanner, Gerhard},
  journal={Acta Numerica},
  volume={12},
  pages={399--450},
  year={2003}
}

@article{verlet1967computer,
  title={Computer ``experiments'' on classical fluids. {I}. {T}hermodynamical properties of {L}ennard-{J}ones molecules},
  author={Verlet, Loup},
  journal={Physical Review},
  volume={159},
  number={1},
  pages={98--103},
  year={1967}
}

@article{benettin1994shadow,
  title={On the {H}amiltonian interpolation of near-to-the-identity symplectic mappings with application to symplectic integration algorithms},
  author={Benettin, Giancarlo and Giorgilli, Antonio},
  journal={Journal of Statistical Physics},
  volume={74},
  number={5--6},
  pages={1117--1143},
  year={1994}
}

@article{mackay1983linear,
  title={Linear stability of periodic orbits in {L}agrangian systems},
  author={MacKay, Robert S. and Meiss, James D.},
  journal={Physics Letters A},
  volume={98},
  number={3},
  pages={92--94},
  year={1983}
}

@article{nesterov1983method,
  title={A method of solving a convex programming problem with convergence rate {$O(1/k^2)$}},
  author={Nesterov, Yurii},
  journal={Soviet Mathematics Doklady},
  volume={27},
  number={2},
  pages={372--376},
  year={1983}
}

@inproceedings{sutskever2013importance,
  title={On the Importance of Initialization and Momentum in Deep Learning},
  author={Sutskever, Ilya and Martens, James and Dahl, George and Hinton, Geoffrey},
  booktitle={International Conference on Machine Learning},
  pages={1139--1147},
  year={2013}
}
\bibliographystyle{abbrvnat}

\appendix

\section{Worked Examples}
\label{app:parabola}

To build intuition, we briefly work through two example loss functions: the quadratic loss and the quartic loss.

\subsection{The Quadratic Loss}

Consider the quadratic loss:
\begin{equation}
L(w) = \frac{1}{2} w^T H w
\end{equation}
The phase-space edge coupling is given as:
\begin{equation}
\mathcal{B}_{\eta, \beta}= \frac{1}{2} w^T H w + \frac{1}{2} w'^T H w' - \frac{1-\beta}{2\eta} \|w - w'\|^2 - \beta (m - m')\cdot (w - w') - \eta \beta m \cdot m'
\end{equation}
And the position-space edge coupling is given as:
\begin{align}
	\mathcal{A}_{\eta,\beta}
	&= \tfrac{1}{2}w^{T}Hw + \tfrac{1}{2}w'^{T}Hw' - \frac{1+\beta}{2\eta}\|w-w'\|^{2},
	\\[4pt]
	&= \tfrac{1}{2}\,w^{T}\!\left(H-\kappa I\right)\!w
	+ \tfrac{1}{2}\,w'^{T}\!\left(H-\kappa I\right)\!w'
	+ \kappa\, w^{T}w'\\[6pt]
	&= \frac{1}{2}
	\begin{bmatrix} w \\ w' \end{bmatrix}^{T}
	\begin{bmatrix}
		H-\kappa I_d & \kappa I_d\\[2pt]
		\kappa I_d & H-\kappa I_d
	\end{bmatrix}  	\begin{bmatrix} w \\ w' \end{bmatrix}\\
	&=
\frac{1}{2}\begin{bmatrix} w \\ w' \end{bmatrix}^{T}\!A \begin{bmatrix} w \\ w' \end{bmatrix}, \quad A = \begin{bmatrix}
	H-\kappa I_d & \kappa I_d\\[2pt]
	\kappa I_d & H-\kappa I_d
\end{bmatrix} 
\end{align}
where $\kappa = \frac{1+\beta}{\eta}$. Note that the matrix $A$ is exactly equal to the Schur complement $S$. 

We can find the set of critical points by taking the derivative of $\mathcal{A}$ and setting it equal to zero:
\begin{equation}
A \begin{bmatrix} w \\ w' \end{bmatrix} = 0 \implies (w, w') \text{ is a fixed point or two-period orbit}.
\end{equation}
We have that when $A$ is non-singular, the only valid solution is $(0_d, 0_d)$---a fixed point, not a two-point orbit. To find a non-trivial two-period orbit, we need that $A$ is singular. Computing the determinant of $A$:
\begin{equation}
\det A = \det (H[ H - 2 \kappa I_d])
\end{equation}
From which it follows:
\begin{equation}
\det A = 0 \implies \det H = 0 \text{ or } \det (H - 2 \kappa I_d) = 0
\end{equation}
The first case corresponds to the Hessian of the loss $H$ having a zero eigenvalue. In that situation, the corresponding eigenvector defines a line of fixed points, since the gradient along that direction vanishes.

The case we are interested in is when $\det(H - 2\kappa I_d) = 0$. This holds if and only if $H$ has an eigenvalue $\lambda$ satisfying
\begin{equation}
	\lambda - 2\kappa = 0 \implies \lambda = 2\kappa = \frac{2(1+\beta)}{\eta},
\end{equation}
which is precisely the sharpness threshold for heavy-ball momentum.

Note that because the existence of two-point orbits requires $\det A = 0$, while stability requires $\det A \neq 0$, there are \textit{no} stable two-period orbits for the quadratic loss. At best they are marginally stable: small perturbations neither expand nor decay, but simply persist.

It is even easier to see the structure of the quadratic loss in centered coordinates. For the centered position-space edge coupling, we have: 
\begin{equation*}
	\Psi(\bar w, \delta) = \bar{w}^T H\,\bar{w} + \delta^T(H - 2\kappa I_d)\,\delta.
\end{equation*}
The centered basis diagonalizes the edge coupling. Assuming that $H$ is positive-definite, we have that:
\begin{equation}
\bar w = 0 \text{  and  } (H - 2\kappa I_d)\delta = 0 \implies (\bar{w}, \delta) \text{  is a fixed point or two-period orbit}.
\end{equation}

\subsection{The Quartic Loss}
\label{app:quartic}

Consider the quartic loss:
\begin{equation*}
	L(w) = \frac{1}{2} H[w,w] - \frac{1}{4} Q[w, w, w, w].
\end{equation*}
For simplicity, we assume that both $H$ and $Q$ are positive-definite. It is important to note that we are now using $H$ to denote the value of the Hessian at the \textit{origin}---for this particular example, it will not denote the Hessian as a generic object which varies throughout space. Throughout, $Q[u,v,z] \in \mathbb{R}^{d}$ and $Q[u,v] \in \mathbb{R}^{d\times d}$ denote partial contractions of the symmetric form $Q$, so that
\begin{equation}
	\nabla L(w) = Hw - Q[w,w,w], \qquad \nabla^{2} L(w) = H - 3\,Q[w,w].
\end{equation}
The phase-space edge coupling is given as:
\begin{equation}
	\begin{aligned}
		\mathcal{B}_{\eta,\beta} ={}& \tfrac{1}{2}H[w,w] - \tfrac{1}{4}Q[w,w,w,w]
		+ \tfrac{1}{2}H[w',w'] - \tfrac{1}{4}Q[w',w',w',w'] \\
		&- \frac{1-\beta}{2\eta}\,\|w-w'\|^{2} - \beta\,(m-m')\cdot(w-w') - \eta\beta\, m\cdot m'
	\end{aligned}
\end{equation}
And the position-space edge coupling is given as:
\begin{align}
	\mathcal{A}_{\eta,\beta}
	&= \tfrac{1}{2}H[w,w] + \tfrac{1}{2}H[w',w']
	- \tfrac{1}{4}Q[w,w,w,w] - \tfrac{1}{4}Q[w',w',w',w']
	- \frac{1+\beta}{2\eta}\,\|w-w'\|^{2}
	\\[4pt]
	&= \frac{1}{2}
	\begin{bmatrix} w \\ w' \end{bmatrix}^{T}\!A
	\begin{bmatrix} w \\ w' \end{bmatrix}
	- \tfrac{1}{4}Q[w,w,w,w] - \tfrac{1}{4}Q[w',w',w',w'],
\end{align}
with $A$ and $\kappa$ having the same definition as before. The set of critical points is given as:
\begin{equation}
	A \begin{bmatrix} w \\ w' \end{bmatrix}
	= \begin{bmatrix} Q[w,w,w] \\ Q[w',w',w'] \end{bmatrix}
	\iff (w, w') \text{ is a fixed point or two-period orbit}.
\end{equation}
There is an important structural difference between the quadratic loss and the quartic loss: nontrivial solutions no longer require $A$ to be singular. The linear part can now be balanced against the cubic term, so nonzero critical points can be isolated and nondegenerate.

We are interested in when there are stable two-point orbits. For the centered position-space edge coupling, we have:
\begin{equation}
	\Psi(\bar w, \delta)
	= H[\bar w, \bar w] + (H - 2\kappa I_d)[\delta, \delta]
	- \tfrac{1}{2}\,Q[\bar w,\bar w,\bar w,\bar w]
	- 3\,Q[\bar w,\bar w,\delta,\delta]
	- \tfrac{1}{2}\,Q[\delta,\delta,\delta,\delta].
\end{equation}
The centered basis no longer diagonalizes the edge coupling: $\bar{w}$ and $\delta$ interact through the cross term $-3\,Q[\bar w,\bar w,\delta,\delta]$.

Consider the case when $\bar{w} = 0$. For the criticality condition for $\delta$, we have:
\begin{equation}
	(H - 2\kappa I_d)\,\delta = Q[\delta,\delta,\delta].
	\label{eq:antipodal-orbit}
\end{equation}
Contracting the above with $\delta$ gives:
\begin{equation}
	\delta^{T}(H - 2\kappa I_d)\,\delta = Q[\delta,\delta,\delta,\delta].
\end{equation}
Recall that $\kappa$ is inversely proportional to the step size $\eta$, so large
$\kappa$ corresponds to small steps. Whenever $\lambda_{\max}(H) < 2\kappa$, the left-hand side is strictly negative for every $\delta \neq 0$. Meanwhile, the right-hand side is nonnegative under our
assumption on $Q$. It would then follow that $\delta = 0$ is the only solution---there are no two-point orbits.

Suppose now that a simple eigenvalue $\lambda$ of $H$ crosses the threshold
$2\kappa$, and let $v$ denote its unit eigenvector. A branch of antipodal orbits emerges
supercritically from the origin as $\lambda$ passes $2\kappa$---with amplitude:
\begin{equation}
	\|\delta\|^{2} = \frac{\lambda - 2\kappa}{Q[v,v,v,v]}
	+ O\!\big((\lambda - 2\kappa)^{2}\big).
\end{equation}
We will now examine its stability. At an antipodal critical point, the Hessian of $\Psi$ is block-diagonal:
\begin{equation}
	\nabla^{2}\Psi(0,\delta)
	= 2\begin{bmatrix} H - 3\,Q[\delta,\delta] & 0 \\ 0 & H - 3\,Q[\delta,\delta] - 2\kappa I_d \end{bmatrix}.
\end{equation}
For our stability condition, we have that:
\begin{equation}
	(0, \delta) \text{ is linearly stable}
	\iff
	\nabla^{2}_{\bar w \bar w}\Psi \succ 0
	\ \text{ and } \
	\nabla^{2}_{\delta\delta}\Psi \prec 0.
\end{equation}
Stable two-period orbits are min--max critical points of the centered edge coupling. They are minima in the midpoint $\bar w$ and maxima in the half-difference $\delta$. 

\section{Nesterov Momentum}
\label{app:nesterov}

This appendix collects the Nesterov analogues of the paper's main objects. 

\subsection{Look-Ahead Coordinates}

First, it helps to change coordinates. Define the \emph{look-ahead point}
\begin{equation}
	\theta_t \;=\; w_t + \eta \beta m_t .
\end{equation}
We have that $\theta_t$ is the momentum-displaced point where the look-ahead gradient is evaluated. Rewriting in the $(\theta, m)$ coordinate system, the update equations become:
\begin{align}
	m_{t+1} &= \beta m_t - \nabla L(\theta_t), \\[2pt]
	\theta_{t+1}
	&= w_{t+1} + \eta \beta m_{t+1}
\end{align}
We will work in look-ahead coordinates throughout. For our phase-space edge coupling, we have:
\begin{equation}
	\mathcal{B}^{\mathrm{NAG}}_{\eta,\beta}
	= L(\theta) + L(\theta') - \frac{1-\beta}{2\eta}\,\|\theta - \theta'\|^2 - \beta^2\,(m - m')\cdot(\theta - \theta') + \frac{\eta\beta^3}{2}\,\|m - m'\|^2 - \eta\beta^2\; m \cdot m'.
	\label{eq:nag-theta-coupling}
\end{equation}
For our momentum criticality conditions, we have:
\begin{align}
	\frac{\partial \mathcal{B}^{\mathrm{NAG}}_{\eta,\beta}}{\partial m}
	&= -\beta^{2}\,(\theta - \theta') + \eta\beta^{3}\,(m - m') - \eta\beta^{2}\, m' \;=\; 0,
	\label{eq:nag-crit-m}\\[4pt]
	\frac{\partial \mathcal{B}^{\mathrm{NAG}}_{\eta,\beta}}{\partial m'}
	&= \phantom{-}\beta^{2}\,(\theta - \theta') - \eta\beta^{3}\,(m - m') - \eta\beta^{2}\, m \;=\; 0.
	\label{eq:nag-crit-mprime}
\end{align}
Dividing through by $\beta^{2}$, these reduce to:
\begin{align*}
	\eta\beta\, m - \eta(1+\beta)\, m' &= \theta - \theta', \\
	\eta(1+\beta)\, m - \eta\beta\, m' &= \theta - \theta'.
\end{align*}
Subtracting the two equations gives $\eta\,(m + m') = 0$---from which it follows that $m' = -m$. Adding the two equations gives
$\eta(1+2\beta)\,(m - m') = 2\,(\theta - \theta')$. Rearranging, we can then derive:
\begin{equation}
	m = \frac{\theta - \theta'}{\eta\,(1+2\beta)}, \qquad m' = -m,
	\label{eq:nag-momentum-solution}
\end{equation}
Substituting our derived expressions for the momenta back in the phase-space edge coupling gives us the position-space edge coupling:
\begin{equation}
	\mathcal{A}^{\mathrm{NAG}}_{\eta,\beta}(\theta, \theta') \coloneqq \operatorname*{stat}_{m, m'}\, \mathcal{B}^{\mathrm{NAG}}_{\eta,\beta}
	= L(\theta) + L(\theta') - \frac{\kappa_\theta}{2}\,\|\theta - \theta'\|^2,
	\qquad \kappa_\theta \coloneqq \frac{1+\beta}{\eta\,(1+2\beta)}.
	\label{eq:nag-A-theta}
\end{equation}

\subsection{The Hessian}
\label{app:nag-hessian}

The second-order partial derivatives of $\mathcal{B}^{\mathrm{NAG}}$ are:
\begin{align*}
\frac{\partial^2 \mathcal{B}^{\mathrm{NAG}}}{\partial \theta^2} &= H_\theta - \frac{1-\beta}{\eta}\, I_d, &
\frac{\partial^2 \mathcal{B}^{\mathrm{NAG}}}{\partial \theta\, \partial \theta'} &= \frac{1-\beta}{\eta}\, I_d, \\
\frac{\partial^2 \mathcal{B}^{\mathrm{NAG}}}{\partial \theta\, \partial m} &= -\beta^2\, I_d, &
\frac{\partial^2 \mathcal{B}^{\mathrm{NAG}}}{\partial \theta\, \partial m'} &= \beta^2\, I_d, \\
\frac{\partial^2 \mathcal{B}^{\mathrm{NAG}}}{\partial m^2} &= \eta\beta^3\, I_d, &
\frac{\partial^2 \mathcal{B}^{\mathrm{NAG}}}{\partial m\, \partial m'} &= -\eta\beta^2(1+\beta)\, I_d,
\end{align*}
where $H_\theta$ denotes the Hessian evaluated at $\theta$. Assembling the blocks in the order $(\theta, \theta', m, m')$:
\begin{equation}
	\nabla^2 \mathcal{B}^{\mathrm{NAG}}_{\eta, \beta} =
	\begin{pmatrix}
		H_\theta - \frac{1-\beta}{\eta} I & \frac{1-\beta}{\eta} I & -\beta^2 I & \beta^2 I \\[3pt]
		\frac{1-\beta}{\eta} I & H_{\theta'} - \frac{1-\beta}{\eta} I & \beta^2 I & -\beta^2 I \\[3pt]
		-\beta^2 I & \beta^2 I & \eta\beta^3 I & -\eta\beta^2(1+\beta) I \\[3pt]
		\beta^2 I & -\beta^2 I & -\eta\beta^2(1+\beta) I & \eta\beta^3 I
	\end{pmatrix}
	\label{eq:nag-hessian-B}
\end{equation}
For the Schur complement of the momentum block, we have:
\begin{equation}
	S^{\mathrm{NAG}} =
	\begin{pmatrix}
		H_\theta - \kappa_\theta\, I & \kappa_\theta\, I \\[3pt]
		\kappa_\theta\, I & H_{\theta'} - \kappa_\theta\, I
	\end{pmatrix}
	= \nabla^2 \mathcal{A}^{\mathrm{NAG}}_{\eta,\beta},
	\label{eq:nag-schur}
\end{equation}
The structure of the LDL factorization is identical to the case of heavy ball. 
For the determinant, we have:
\begin{equation}
	\det \nabla^2 \mathcal{B}^{\mathrm{NAG}}_{\eta,\beta} = \bigl(-\eta^2\beta^4(1+2\beta)\bigr)^d\, \det S^{\mathrm{NAG}}_{\eta,\beta} = \beta^{4d}\, \det\bigl(M^{\mathrm{NAG}} - I\bigr),
	\label{eq:nag-det-chain}
\end{equation}
where $M^{\mathrm{NAG}}$ is the two-step Jacobian of Nesterov momentum.

\section{Deferred Proofs}

In this section, we will prove various key assertions made throughout the paper.

\subsection{Equivalence of the Phase-Space and Position-Space Pencils}
\label{app:pencil-proof}

Let $P$ denote the linear pencil on phase space and $W$ denote the quadratic pencil on position space:
\begin{equation}
	P(\lambda) \coloneqq M - \lambda I_{2d}, \qquad
	W(\lambda) \coloneqq (\lambda + \beta)^2 I_d - \lambda A_{w'} A_w .
\end{equation}
We will first show that the momentum variables can be eliminated from the linear eigenvalue equation for $M$, leaving a position-space eigenvalue equation governed by $W$. We will then prove the stronger statement: the determinants of the two pencils agree at \textit{every} value of $\lambda$, not just at their zeros.

Let $\xi = (x, p)$ be an eigenvector of $M$ with eigenvalue $\lambda$, and let $\zeta = (y, q) = J(H_w)\, \xi$ denote its image under the first half of the period: $J(H_{w'})\, \zeta = \lambda \xi$. Writing out the components of the two half-period maps:
\begin{align*}
	\zeta &= J(H_w) \xi: &\quad q &= \beta p - H_w x, & y &= x + \eta q, \\
	\lambda \xi &=  J(H_w') \zeta: &\quad \lambda p &= \beta q - H_{w'} y, & \lambda x &= y + \eta \lambda p .
\end{align*}
Using the rightmost column, we can express both momenta in terms of the positions: 
\begin{equation*}
 q = \tfrac{1}{\eta} (y - x), \qquad  p = \tfrac{1}{\eta}(x - y).
\end{equation*} 
Substituting into the two equations in the middle column and collecting terms, we obtain a pair of equations expressed solely using positions:
\begin{equation}
	(\lambda + \beta)\, y = \lambda A_w\, x, \qquad
	(\lambda + \beta)\, x = A_{w'}\, y .
	\label{eq:half-period}
\end{equation}
where $A_w = \eta (\kappa I_d - H_w)$. Eliminating the intermediate position $y$, we obtain:
\begin{equation*}
	(\lambda + \beta)^2\, x = \lambda A_{w'} A_w\, x \iff W(\lambda)\, x = 0,
\end{equation*}
which is precisely the position-space eigenvalue equation.

We will now prove the determinant identity:
\begin{equation}
\det P(\lambda) = \det W(\lambda)
\end{equation}
Recall that the two-step Jacobian is given by the product of the one-step Jacobians: $M = J(H_{w'} ) J(H_w)$. Multiplying out the one-step Jacobians, we have:
\begin{equation}
	M = \begin{pmatrix}
		A_{w'} A_w - \beta A_{w'} - \beta I_d & \eta\beta\, A_{w'} \\[3pt]
		\frac{1}{\eta}\bigl(A_{w'} A_w - A_w - \beta A_{w'}\bigr) & \beta A_{w'} - \beta I_d
	\end{pmatrix}.
\end{equation}
Consider the linear pencil $P(\lambda)$. Adding $-\eta$ times the momentum row of
$P(\lambda)$ to its position row leaves the determinant unchanged, since the
operation amounts to left-multiplication by a unit triangular matrix. We can
then deduce
\begin{align}
	\det P(\lambda)
	&= \det\bigl(M - \lambda I_{2d}\bigr) \notag \\[3pt]
	&= \det
	\begin{pmatrix}
		A_w - (\lambda+\beta) I
		& \eta(\lambda+\beta)\, I \\[3pt]
		\dfrac{1}{\eta}\bigl(A_{w'} A_w - A_w - \beta A_{w'}\bigr)
		& \beta A_{w'} - (\lambda+\beta) I
	\end{pmatrix}.
	\label{eq:P-swept}
\end{align}
The upper-right block is now a multiple of the identity. We can then transpose the top and bottom rows so that the two blocks of the bottom row commute---so we have that our block-determinant identity now applies:
\begin{equation*}
	\det P(\lambda) = \det\Bigl(\bigl[\beta A_{w'} - (\lambda+\beta) I\bigr]\bigl[A_w - (\lambda+\beta) I\bigr] - (\lambda+\beta)\bigl[A_{w'} A_w - A_w - \beta A_{w'}\bigr]\Bigr).
\end{equation*}
When we expand, every cross term cancels. We can then deduce:
\begin{equation}
	\det P(\lambda) = \det\bigl((\lambda+\beta)^2 I_d - \lambda A_{w'} A_w\bigr) = \det W(\lambda)
	\label{eq:pencil-identity}
\end{equation}
which holds for all $\lambda$.

We can also prove the identity of \cref{eq:pencil-det}. For $\lambda > 0$, we have:
\begin{equation*}
	W(\lambda) = -\lambda\eta^2 \Bigl[(\kappa I - H_{w'})(\kappa I - H_w) - \nu_\lambda^2\, I\Bigr],
	\qquad \nu_\lambda^2 = \frac{(\lambda+\beta)^2}{\eta^2 \lambda}.
\end{equation*}
The bottom-row blocks of $S(\lambda)$ commute, so the block-determinant identity gives $\det S(\lambda) = \det\bigl[(H_w - \kappa I)(H_{w'} - \kappa I) - \nu_\lambda^2 I\bigr]$. Even though the two products appear in opposite order in their respective determinants, we have that the two matrices are transposes of one another---so their determinants agree. It then follows that:
\begin{equation*}
	\det P(\lambda) = (-1)^d \lambda^d \eta^{2d} \det S(\lambda),
\end{equation*}
which is precisely \cref{eq:pencil-det}.

\subsection{Critical Points of the Two-Period Phase-Space Action}
\label{app:action-proof}

We will now verify that the critical positions of the phase-space edge coupling and the two-period action agree, while their critical momenta differ by a constant scaling factor. Throughout, we will have that $\Delta t = \sqrt{2\eta/(1+\beta)}$.

Stationarity of $\mathcal{S}^{(2)}_{\mathrm{ph}, d}$ with respect to the two momenta gives:
\begin{equation*}
	\frac{\partial \mathcal{S}^{(2)}_{\mathrm{ph}, d}}{\partial m} = (w - w') - \Delta t\, m = 0, \qquad
	\frac{\partial \mathcal{S}^{(2)}_{\mathrm{ph}, d}}{\partial m'} = -(w - w') - \Delta t\, m' = 0.
\end{equation*}
From which we can deduce:
\begin{equation}
	m = \frac{w - w'}{\Delta t}, \qquad m' = -m .
	\label{eq:action-momenta}
\end{equation}
Stationarity with respect to the two positions gives:
\begin{equation*}
	m - m' = \Delta t\, \nabla L(w), \qquad -(m - m') = \Delta t\, \nabla L(w') .
\end{equation*}
Substituting the momenta, we can obtain two equations expressed solely in terms of the positions:
\begin{equation}
	\nabla L(w) = \frac{2}{\Delta t^2}\,(w - w'), \qquad
	\nabla L(w') = \frac{2}{\Delta t^2}\,(w' - w) .
	\label{eq:action-positions}
\end{equation}
At $\Delta t = \sqrt{2\eta/(1+\beta)}$, we have that $2/\Delta t^2 = \kappa$.

Now consider the edge coupling. Criticality of the momentum of $\mathcal{B}_{\eta,\beta}$ gives us:
\begin{equation} 
\eta m = w - w', \quad m' = -m. 
\end{equation}
Substituting these into the momentum update $m' = \beta m - \nabla L(w)$ yields:
\begin{equation*}
	\nabla L(w) = \kappa\,(w - w').
\end{equation*}
By the $(z \leftrightarrow z')$ symmetry, we also have that $\nabla L(w') = \kappa\,(w' - w)$. These are exactly the equations in \cref{eq:action-positions}: the two functionals impose the \textit{same} conditions on $(w, w')$, so their critical positions coincide.

The critical momenta do not coincide. Both point along $w - w'$, but the action carries momenta $(w - w')/\Delta t$ whereas the edge coupling carries momenta $(w - w')/\eta$. The critical momenta of the action are therefore obtained from those of the edge coupling by the fixed rescaling $\eta/\Delta t = \sqrt{\eta(1+\beta)/2}$.

\end{document}